\documentclass[journal]{IEEEtran}
\pdfoutput=1
\ifCLASSINFOpdf
\else
   \usepackage[dvips]{graphicx}
\fi
\usepackage{url}

\usepackage{graphicx}
\usepackage{lineno}
\usepackage{amsmath}
\DeclareMathOperator*{\argmin}{arg\,min}
\usepackage{diagbox}
\usepackage{multirow}
\usepackage{colortbl}

\definecolor{gbypink}{rgb}{0.99, 0.91, 0.95}
\definecolor{gbygreen}{rgb}{0.95, 0.99, 0.91}
\definecolor{gbyblue}{rgb}{0.91, 0.95, 0.99}

\begin{document}

\title{Mobile Image Restoration via Prior Quantization}

\author{Shiqi Chen, Jinwen Zhou, Menghao Li, Yueting Chen, Tingting Jiang
\thanks{This work was supported in part by the Civil Aerospace Pre-Research Project D040107, the National Natural Science Foundation of China 61424080211, and the Key Research Project of Zhejiang Lab 2021MH0AC01. (Corresponding author: Tingting Jiang.)}
\thanks{Shiqi Chen, Jinwen Zhou, and Yueting Chen are with the College of Optical Science and Engineering, Zhejiang University, Hangzhou 310000, China (e-mail: chenshiqi@zju.edu.cn).}
\thanks{Tingting Jiang is with the Research Center for Intelligent Sensing Systems, Zhejiang Laboratory, Hangzhou 311100, China (e-mail: eagerjtt@zhejianglab.com).}}

\markboth{Journal of \LaTeX\ Class Files, Vol. 14, No. 8, May 2023}
{Shell \MakeLowercase{\textit{et al.}}: Bare Demo of IEEEtran.cls for IEEE Journals}
\maketitle

\begin{abstract}
In digital images, the performance of optical aberration is a multivariate degradation, where the spectral of the scene, the lens imperfections, and the field of view together contribute to the results. Besides eliminating it at the hardware level, the post-processing system, which utilizes various prior information, is significant for correction. However, due to the content differences among priors, the pipeline that aligns these factors shows limited efficiency and unoptimized restoration. Here, we propose a prior quantization model to correct the optical aberrations in image processing systems. To integrate these messages, we encode various priors into a latent space and quantify them by the learnable codebooks. After quantization, the prior codes are fused with the image restoration branch to realize targeted optical aberration correction. Comprehensive experiments demonstrate the flexibility of the proposed method and validate its potential to accomplish targeted restoration for a specific camera. Furthermore, our model promises to analyze the correlation between the various priors and the optical aberration of devices, which is helpful for joint soft-hardware design.
\end{abstract}


\IEEEpeerreviewmaketitle

\section{Introduction}

\IEEEPARstart{A}{ny} digital imaging system suffers from optical aberration, and thus correcting aberration is necessary for accurate measurements \cite{yue2015blind}. Unfortunately, the optical aberration in image is affected by many factors, which is formulated as:
\begin{equation}
    \label{eq:image formation model}
    J_{e} = \int \mathcal{C}_{e}(\lambda)\cdot [I_{e}(h,w,\lambda)*L_{e}(h,w,\lambda)]d\lambda + N_{e}(h, w),
\end{equation}
here $(h,w)$ indicates the coordinates of the pixel, $\lambda$ is the wavelength, $\mathcal{C}_{e}$ and $I_{e}$ denote the spectral response and the point spread function (PSF), $L_{e}$, $J_{e}$, and $N_{e}$ are the latent sharp image, the observed image, and the measurement noise, respectively. Note that the subscript $e$ indicates the measurement that represents the energy received by the sensor. Therefore, energy dispersion and field-of-view (FoV) clue of the optical system, spectral properties, and sensor noise together contribute to the optical aberration expression on digital images. In other words, these factors serve as the priors to facilitate the correction.

Although there are many algorithms for optical aberration correction, it still faces a few challenges for widespread application. One issue is that the existing methods generally have a limited application scope, \textit{e.g.}, the deconvolution is inefficient in handling spatial-varying kernels, and the deep learning method is trained for a specific device according to the data \cite{eboli2022fast}. Another challenge is the quality of restoration, which is generally unsatisfactory due to the lack of sufficient information. An inherent defect of these unflexible methods is that they are incapable of mining the interaction between the digital pixel and other optical priors \cite{9864277}.

As mentioned above, the expression of optical aberration correlates with multiple factors. Designing a general method to utilize these priors for targeted correction is an issue worth discussing. Meanwhile, this baseline can help analyze the correlation between different priors and optical aberration correction, aiming to guide the co-design of the hardware configuration and the post-processing pipeline in the high-end imaging system.

In this letter, we propose a prior quantization model to correct the optical degradation influenced by multiple factors, where different priors corresponding to each factor are fed to the model. However, due to the content differences among the multimodal priors, they cannot directly integrate into the model for efficient post-processing. To this end, we encode the various priors into a high-dimensional latent space and characterize it by a learnable codebook. The learned code reduces the dimension of multimodal priors representation and models the global interrelations of auxiliary information. Then the model integrates the quantized prior representation into the image restoration branch by fusing it with features of different scales. Finally, we supervise the restoration in multiple scales to ensure that the cross-scale information used for fusion is accurate and valid. Instead of a black box, the model can adjust the input pixel-level prior according to the demands of the user, aiming to achieve a targeted development. Moreover, the learned codebook bridges the gap between different priors and restoration. Therefore, the proposed model has the potential to analyze the utilization of priors, which is meaningful for imaging system design.

\section{The Proposed Method}

% \subsection{Prior Quantization Model}

\begin{figure*}[ht]
    \centering
    \includegraphics[width=\linewidth]{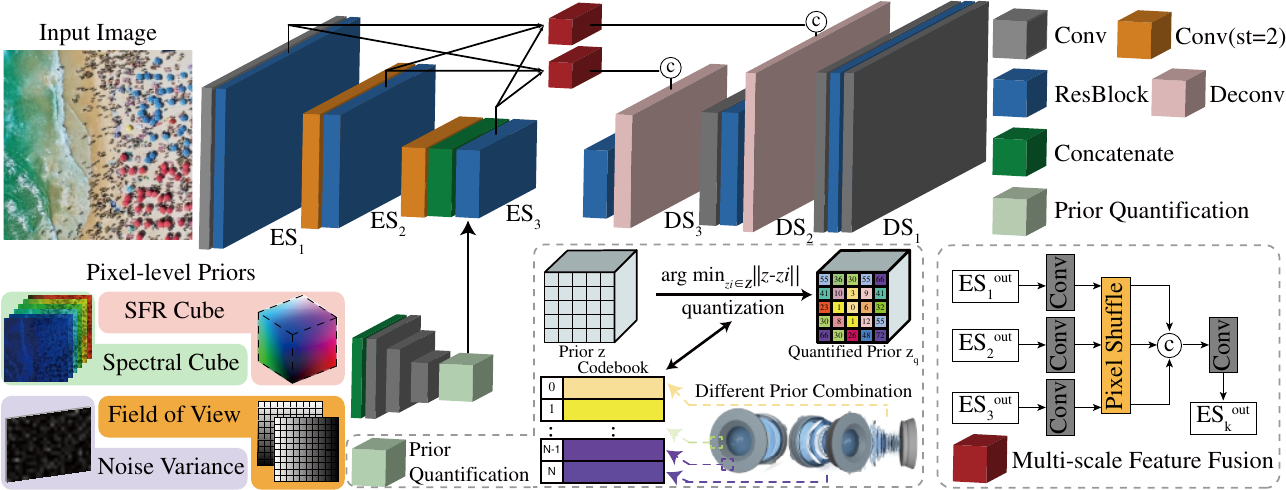}
    \caption{The detail of the proposed prior quantization model. The layer configurations are illustrated with different colored blocks.}
    \label{fig:network architecture}
\end{figure*}

As mentioned above, multiple factors contribute to the expression of optical aberration. Thus the model must possess the ability to perceive a complex combination and transform it into a simpler form. Directly engaging the image signal with individual priors is inefficient, where the model will spend substantial computing overhead on the auxiliary information. Therefore, we encode these priors into a high-dimension space and represent them with a learnable codebook. Since the aberration is relatively constant in the neighborhood, the optical prior of an image $x \in \mathcal{R}^{H \times W \times 3}$ can be represented by a spatial collection of codebook $z_{q} \in \mathcal{R}^{h \times w \times d_{z}}$, where $d_{z}$ is the dimension of the code. To effectively learn such a codebook, we propose to exploit the spatial extraction capability of convolution and incorporate with the neural discrete representation learning \cite{van2017neural}. First, we use CNNs to encode the priors into a latent space with the same spatial resolution ($h \times w$). Then, we embed the encoded features into the spatial code $\hat{z}_{ij} \in \mathcal{R}^{d_{z}}$ and quantify each code onto its closest entry in the discrete codebook $\mathcal{Z}=\{z_{k}\}^{K}_{k=1}\subset\mathcal{R}^{n_{z}}$:
\begin{equation}
    \label{eq:quantization}
    z_{q} = \left(\argmin_{z_{k} \in \mathcal{Z}} ||\hat{z}_{ij}-z_{k}||\right)\in\mathcal{R}^{h \times w \times d_{z}},
\end{equation}

Third, we concatenate the quantized prior representation with the features in the restoration branch. And the subsequent ResBlock actively fused the image feature, where the prior after quantization provides clues of the spatial/channel importance to the reconstruction branch. 

In most reconstruction models, the skip connections are only processed in one scale when the coarse-to-fine architecture is applied. Inspired by the dense connection between multi-scale features, we implement a multi-scale feature fusion (MFF) module to fuse the quantified representation with the features from other scales \cite{cho2021rethinking}. As shown in the right-bottom of Fig. \ref{fig:network architecture}, this module receives the outputs of different encoding scales ($ES_{i}$) as inputs. After adjusting the channels of each $ES^{out}_{i}$ by convolution, we use pixel shuffle to transform the encoded information into the same spatial resolution and then perform the fusion \cite{shi2016real}. The output of the MFF is delivered to its corresponding decoding scales. 
% More specifically, the MFFs of different scales are formulated as follows:
% \begin{equation}
%     \begin{aligned}
%         MFF_{1} &= \mathcal{C}_{DS1}(\{\mathcal{P}(\mathcal{C}_{\times1}(ES_{1})), \mathcal{P}(\mathcal{C}_{\times2}(ES_{2})), \mathcal{P}(\mathcal{C}_{\times4}(ES_{3}))\}), \\
%         MFF_{2} &= \mathcal{C}_{DS2}(\{\mathcal{P}(\mathcal{C}_{\times0.5}(ES_{1})), \mathcal{P}(\mathcal{C}_{\times1}(ES_{2})), \mathcal{P}(\mathcal{C}_{\times2}(ES_{3}))\}),
%     \end{aligned}
% \end{equation}
% where $\mathcal{C}$ is the operation of convolution, and the subscript refers to the ratio by which the number of input channels is to be multiplied after convolution. $\mathcal{P}$ is the pixel shuffle to rearrange the features of different scales into the same spatial resolution so that they be concatenated (denoted by $\{\cdot\}$ operation) together. 
In this way, each scale can perceive the encoded information of other scales, especially the lowest-scale features filtered by the quantified priors, resulting in improved restoration quality.

\begin{table*}[ht]
    \center{
        \caption{Performance of the proposed model and other competing methods on synthetic and real data. The percentage denotes the relative improvement compared with the best models (PSNR first turn to RMSE, other metrics have no change).}
        \label{tab:indicator contrast}
        \footnotesize
        \begin{tabular}{|c|r r|r r|r r|r r|r r|r r|}
            \hline
            \multirow{2}{*}{\diagbox[dir=SE]{Method}{TestSet}} & \multicolumn{8}{c|}{Synthetic Evaluation} & \multicolumn{4}{c|}{Real Evaluation} \\
            \cline{2-13}
            & \multicolumn{2}{c|}{PSNR$\uparrow$}  & \multicolumn{2}{c|}{SSIM$\uparrow$} & \multicolumn{2}{c|}{VIF$\uparrow$} & \multicolumn{2}{c|}{LPIPS$\downarrow$} & \multicolumn{2}{c|}{BRISQUE$\downarrow$} & \multicolumn{2}{c|}{NIQE$\downarrow$} \\
                                   
            \hline
            SRN                     & 29.79 & (56.6\%) & 0.9247 & (29.1\%) & 0.8221 & (11.3\%) & 3.428  & (48.2\%) & 54.49 & (16.1\%) & 5.686 & (21.3\%)\\
            IRCNN                   & 30.58 & (48.0\%) & 0.9289 & (24.5\%) & 0.8314 & (10.0\%) & 3.151  & (43.7\%) & 53.47 & (14.5\%) & 5.572 & (19.7\%)\\
            Self-Deblur             & 32.23 & (24.0\%) & 0.9353 & (17.5\%) & 0.8544 & (7.05\%) & 2.766  & (35.8\%) & 50.28 & (9.03\%) & 5.323 & (16.0\%)\\
            GLRA                  & \cellcolor{gbyblue} 32.47 & \cellcolor{gbyblue} (19.6\%)  & 0.9347 & (18.2\%) & 0.8662 & (5.60\%) & 2.457  & (27.8\%) & 49.74 & (8.04\%) & 5.476 & (18.3\%)\\
            FDN                     & \cellcolor{gbygreen} 33.27 & \cellcolor{gbygreen} (3.39\%)  & \cellcolor{gbygreen} 0.9402 & \cellcolor{gbygreen} (12.2\%) & \cellcolor{gbyblue} 0.8943 & \cellcolor{gbyblue} (2.28\%) & \cellcolor{gbygreen} 1.944  & \cellcolor{gbygreen} (8.69\%) & \cellcolor{gbyblue} 47.42 & \cellcolor{gbyblue} (3.54\%) & \cellcolor{gbyblue} 5.038 & \cellcolor{gbyblue} (11.2\%)\\
            Deep Wiener             & 32.02 & (27.6\%)  & \cellcolor{gbyblue} 0.9463 & \cellcolor{gbyblue} (5.7\%) & \cellcolor{gbygreen} 0.9007 & \cellcolor{gbygreen} (1.55\%) & \cellcolor{gbyblue} 1.928  & \cellcolor{gbyblue} (7.94\%) & \cellcolor{gbygreen} 46.52 & \cellcolor{gbygreen} (16.8\%) & \cellcolor{gbygreen} 4.943 & \cellcolor{gbygreen} (9.44\%)\\
            Ours                    & \cellcolor{gbypink} 33.42 & \cellcolor{gbypink} (0.0\%)  & \cellcolor{gbypink} 0.9517 & \cellcolor{gbypink} (0.0\%) & \cellcolor{gbypink} 0.9147 & \cellcolor{gbypink} (0.0\%) & \cellcolor{gbypink} 1.775  & \cellcolor{gbypink} (0.0\%) & \cellcolor{gbypink} 45.74 & \cellcolor{gbypink} (0.0\%) & \cellcolor{gbypink} 4.476 & \cellcolor{gbypink} (0.0\%)\\
            \hline
    \end{tabular}}
\end{table*}
 
Here we discuss the loss function of our model. Due to the quantization operation in Eq. \ref{eq:quantization} is non-differentiable in backpropagation, the CNN encoder of priors cannot receive a gradient to optimize its parameters if the entire network is directly trained with end-to-end supervision. Fortunately, the strategy of the codebook alignment allows the gradient propagated from decoders to update these CNN encoders. Specifically, we keep the encoded priors approaching the vectors of the learnable codebook. In this way, the quantization progress like a gradient estimator, allowing the CNN encoder to estimate the backpropagation and update the parameters even when the gradient is truncated in training. Thus, we apply the codebook alignment loss to realize the optimization of encoders \cite{duan2022multi}:
\begin{equation}
    \label{eq:codebook aligning}
    \mathcal{L}_{align} = ||sg[\hat{z}_{ij}] - z_{q}||_{2}^{2} + ||\hat{z}_{ij} - sg[z_{q}]||_{2}^{2},
\end{equation}
here $\hat{z}_{ij} = \{E_{1}(p_{1}), E_{2}(p_{2}), \dots, E_{n}(p_{n})\}$, where $p_{i}$ and $E_{i}$ are the $i^{th}$ prior and encoder, respectively. $\{\cdot\}$ is the operation to concatenate the encoded features. $sg[\cdot]$ denotes the stop-gradient operation to ensure the loss function only guide the encoded priors and the codebook vectors approaching.

Because our network relies on the refinement in a coarse-to-fine manner, we supervise the image reconstruction on different scales. For the supervision of image content, we find that the L1 loss performs better on quantitative metrics for optical aberration correction. However, since the content loss only measures the pixel-level difference, the model does a good job restoring the low-frequency information (such as brightness and color, \textit{etc.}) while performing poorly in restoring the textures of the scene. To prevent the limitation of pixel-level supervision, we adopt the supervision in the Fourier domain as an auxiliary loss \cite{cho2021rethinking}. Compared with the gradient constraints (\textit{e.g.}, total variation) or the feature similarity on the pre-trained model (\textit{e.g.}, perceptual loss), this item provides more information on different spatial frequencies. The loss $\mathcal{L}_{content}$ is formulated as follows:
\begin{equation}
    \label{eq:content loss}
    \mathcal{L}_{content} = \frac{1}{t_{k}}\sum^{K}_{k=1}||\hat{I_{k}} - I_{k}||_{1} + \lambda ||\mathcal{F}(\hat{I_{k}}) - \mathcal{F}(I_{k})||_{1},
\end{equation}
here the output of the k-th scale is $I_{k}$ and its corresponding ground-truth is $\hat{I_{k}}$, where $K$ is the number of scales. The content loss in each scale is the average on the total elements, whose number is denoted as $t_{k}$. $\mathcal{F}$ is the fast Fourier transform (FFT) operation that performs on different scales and the $\lambda$ is experimentally set to 0.1. The overall loss function is the sum of $\mathcal{L}_{content}$ and $\mathcal{L}_{align}$.

\subsection{Synthetic Flow for training data and prior}
\begin{figure}[t]
    \centering
    \includegraphics[width=\linewidth]{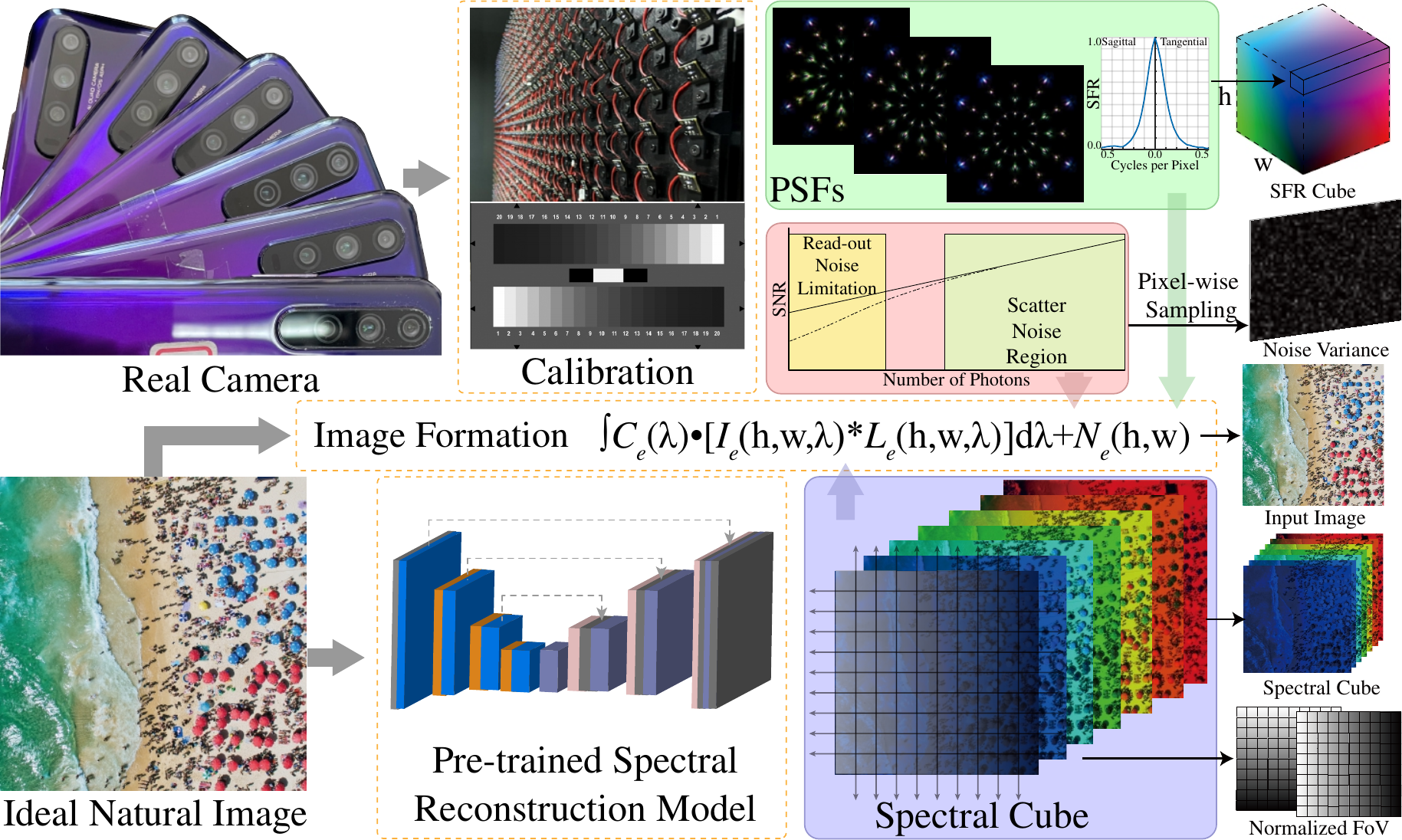}
    \caption{The synthetic flow of the training data and the priors.}
    \label{fig:flow}
\end{figure}

Since optical aberration is highly related to multiple factors, we construct a comprehensive dataset for correction. The detailed synthetic flow of the training data and the priors is shown in Fig. \ref{fig:flow}. First, we calibrate the PSFs and the noise factors of many mobile cameras, where the PSFs are used to simulate the degradation and calculate the SFR prior. Second, we use the pre-trained MST++ \cite{cai2022mst++} to convert RGB images to hyperspectral data for the spectral prior. Third, the FoV prior is consist of the $(h, w)$ pixel coordinates that normalized to $[-1, 1]$ \cite{chen2021extreme,10.1145/3450626.3459674}. Finally, the multispectral data is engaged with optical aberration, where the procedure is the same as Eq. \ref{eq:image formation model}.

\section{Experimental Results}
\label{sec:experiments}

\subsection{Datasets, Metrics, and Training Settings}
As illustrated above, we use the natural images in DIV8K to synthetic the training data-pairs and priors \cite{gu2019div8k}. The training dataset consists of 800 image pairs. And the resolution of image is rescaled to $3000 \times 4000$ to align with the real cameras. In the case of real-world evaluation, we capture many photographs with multiple mobile terminal of Huawei Honor 20 and iPhone 12. As for the metrics, the PSNR, SSIM, VIF \cite{sheikh2006image}, and LPIPS \cite{zhang2018unreasonable} evaluate the model with reference. The BRISQUE \cite{mittal2012no}, NIQE \cite{mittal2012making} are used to assess the restoration of the captrued photographs. In the training, we crop the whole image to $256 \times 256$ pixels and form minibatches of 8 images. For a fair comparisons, the priors are concatnated with the images and fed into the competing models. We optimize the model with Adam in $\beta_{1}=0.5$ and $\beta_{2}=0.999$. The initial learning rate is $10^{-4}$ and then halved every 50 epoch.

\subsection{Quantitative Assessment to SOTA Methods}

We compare the proposed prior quantization model with the competing algorithms designed for optical aberration correction. All these methods are retrained with the same dataset until convergence. In the assessment of spatial-various aberration correction, we evaluate the quantitive indicators on the synthetic dataset. Tab. \ref{tab:indicator contrast} reports the performance of various approaches on the synthetic dataset. The blind methods (SRN \cite{tao2018scale}, IRCNN\cite{zhang2017learning}, SelfDeblur\cite{ren2020neural}, GLRA\cite{ren2018deep}) successfully deal with the optical degradation of one camera. But there are various mobile cameras with different optical aberrations, and the blind manner fails to acquire the precise degradation clue only from the feature of the image. The non-blind approaches (FDN\cite{kruse2017learning}, Deep Wiener\cite{Lin:22}), which feed the pre-trained models with PSFs to adapt to a specific camera, have a similar idea to ours. However, since the PSFs is a high-dimensional representation of degradation, only a few representative PSFs are selected to be fed into the model (only 5 PSFs in \cite{Lin:22}), where the spatial relationship of the PSFs across the whole FoV is abandoned. Otherwise, the complexity will be extremely high. Different from this method, we use the rearranged SFR cube to represent the dispersion for each FoV, which is a lower-dimensional representation of degradation. Therefore, with relatively lower computational overhead, our model can obtain pixel-level guidance to achieve better restoration.

\subsection{Real Restoration Comparisons}
\begin{figure*}[t]
    \centering
    \includegraphics[width=\linewidth]{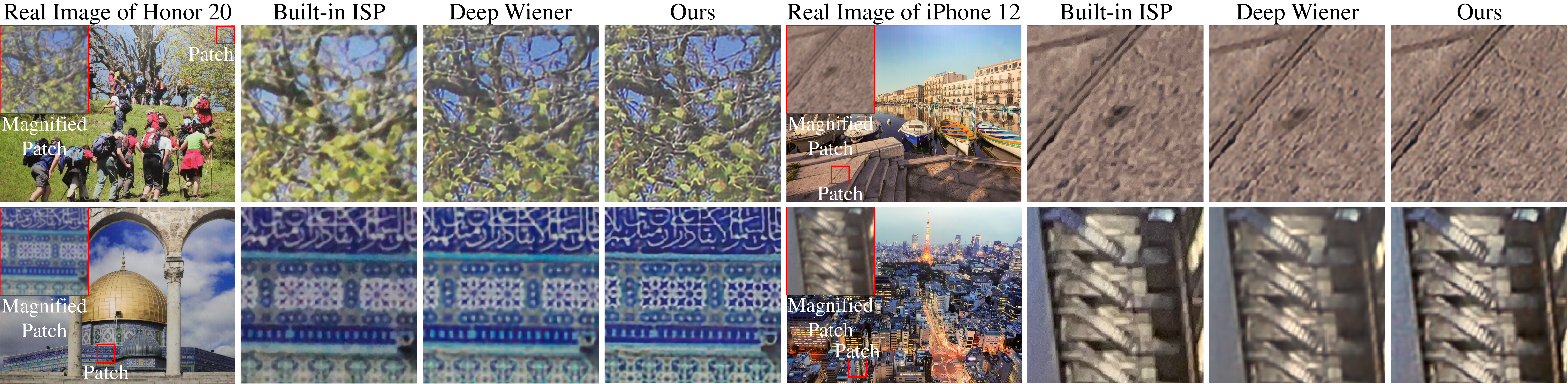}
    \caption{Real image restoration comparison, where the position is highlighted in red. See more experimental results in Supplement 1.}
    \label{fig:real image restoration}
\end{figure*}
Moreover, we test these algorithms with the degraded photographs taken from real cameras, and the results are visualized in Fig. \ref{fig:real image restoration}. Since the priors acquired by the Deep Wiener do not correspond to the pixel neighborhood, the model needs attention mechanism to determine the specific degradation of the input. These overheads increase the processing burden, resulting in the failure to obtain efficient restoration. Compared with other algorithms and the built-in ISP, our model efficiently integrates various priors and characterizes them with the quantized codebook vectors. Therefore, our method reduces the complexity of multi-task post-processing, and achieves a comprehensive improvement in image quality. Another advantage of our non-blind model is that it can be designed as a post-processing system with better generalization. In our experiment, we use the model trained on the simulation dataset to post-process the real images taken by various mobile cameras. The results in the left and right of Fig. \ref{fig:real image restoration} demonstrate that the model pre-trained on synthetic data achieves perfect generalization ability on specific devices (Honor 20 pro and iPhone 12), where fine-tuning is not necessary. Therefore, the plug-and-play feature indicates that the learned model is promising to replace the camera-specific ISP system with a flexible model. For the detailed comparisons, we refer the readers to the supplementary for more restorations on different manufactured samples and more the quantitive results of system imaging quality. 

\subsection{Prior Correlation Analyse}
\begin{figure}[t]
    \centering
    \includegraphics[width=\linewidth]{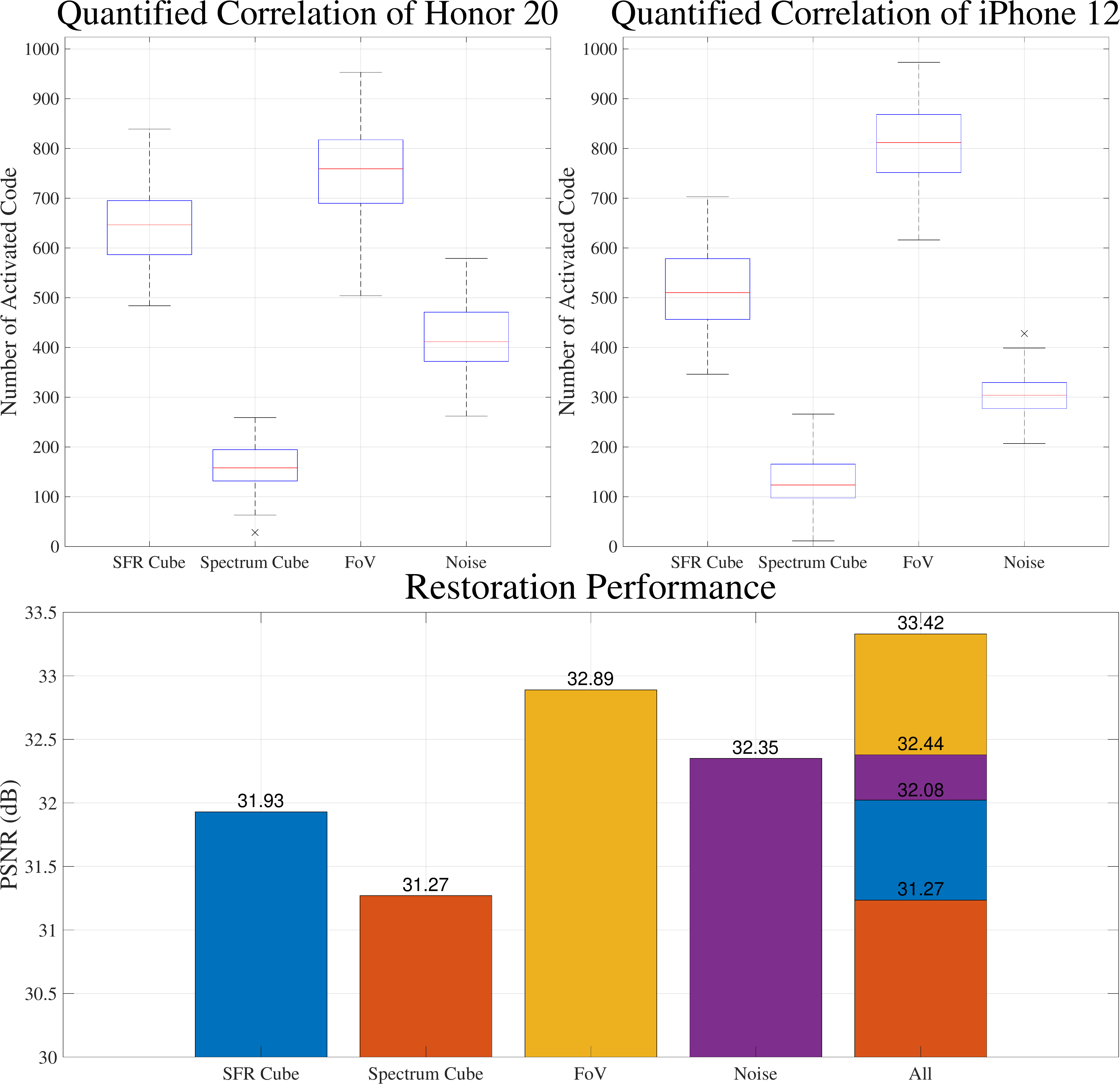}
    \caption{The correlation between priors and optical aberration correction in different mobile cameras.}
    \label{fig:prior correlation}
\end{figure}
As mentioned above, the optical aberration is explicitly correlated to multiple factors when imaging at a fixed distance and this auxiliary information can benefit the restoration. However, obtaining so much assistance in real-time imaging comes at an unaffordable expense. We use the learned model into analyze the proceeds of introducing each prior to the restoration. When evaluating a specific prior, we zero out all other auxiliaries except this one and test the learned model with the raw photographs taken by various mobile cameras, counting the number of the activated codes in inference process. The larger number of activated codebooks indicates the more relevance between this prior and optical aberration. The result of the assessment is shown at the top of Fig. \ref{fig:prior correlation}. We find that the SFR and FoV priors will activate more codes in inference, which means they are highly correlated with aberration and play critical roles in correction. We note that the restoration of the Honor 20 pro activates more code in the SFR and noise priors evaluation when compared with the iPhone 12, which attributes to its uneven optical aberration of the optics and lower SNR of the sensor. This experiment also demonstrates that the number of activation hints at the utilization of different priors, providing a brand new issue for imaging system assessment. Moreover, we evaluate the correlation between the restoration indicators and the priors on the synthetic testset (shown at the bottom of Fig. \ref{fig:prior correlation}). Different from the number of activated codes, we note that the noise is more substantial on the restoration performance, which may be put down to the PSNR that calculates the absolute error in the pixel scale. On the other hand, this phenomenon indicates that we can strip the noise and the optical aberration to balance the model with restoration indicators and computational efficiency.

\section{Conclusion}
In summary, we develop a deep learning model to utilize multiple priors for optical aberration correction. Even though the content of the priors varies a lot from each other, the proposed quantization strategy efficiently fuses these factors and guides the correction of optical aberration. Comprehensive experiments show that integrating all the related information does benefit the restoration of optical aberration, and the proposed model can generalize to different mobile devices without finetuning. Moreover, the learned model quantifies the correlation and significance of different priors for the correction procedure. Thus, the aberration correction in the post-processing system can be more efficient when the critical clues are obtained. In the future, we will put efforts into engaging the proposed model with different imaging devices, aiming to realize efficient correction and better generalization.

\bibliographystyle{IEEEtran}
\bibliography{IEEEabrv, main}

\end{document}